\pdfoutput=1

\documentclass[11pt]{article}

\usepackage{acl}

\usepackage{times}
\usepackage{latexsym}
\usepackage{makecell}
\usepackage{amssymb}
\definecolor{mydarkgreen}{RGB}{0,165,0}
\definecolor{myorange}{RGB}{255,100,0}
\usepackage[ruled,vlined]{algorithm2e}
\usepackage{hyperref}

\usepackage[T1]{fontenc}

\usepackage[utf8]{inputenc}

\usepackage{microtype}

\usepackage{inconsolata}
\usepackage{graphicx}

\title{OpenFMNav: Towards Open-Set Zero-Shot Object Navigation via Vision-Language Foundation Models}

\author{Yuxuan Kuang \\
  Peking University\\
  \texttt{kuangyuxuan@stu.pku.edu.cn} \\\And
  Hai Lin \\
  University of Notre Dame \\
  \texttt{hlin1@nd.edu} \\\And
  Meng Jiang \\
  University of Notre Dame \\
  \texttt{mjiang2@nd.edu}
}

\begin{document}
\maketitle
\begin{abstract}
Object navigation (ObjectNav) requires an agent to navigate through unseen environments to find queried objects. Many previous methods attempted to solve this task by relying on supervised or reinforcement learning, where they are trained on limited household datasets with close-set objects. However, two key challenges are unsolved: understanding free-form natural language instructions that demand open-set objects, and generalizing to new environments in a zero-shot manner. Aiming to solve the two challenges, in this paper, we propose \textbf{OpenFMNav}, an \textbf{Open}-set \textbf{F}oundation \textbf{M}odel based framework for zero-shot object \textbf{Nav}igation. We first unleash the reasoning abilities of large language models (LLMs) to extract proposed objects from natural language instructions that meet the user's demand. We then leverage the generalizability of large vision language models (VLMs) to actively discover and detect candidate objects from the scene, building a \textit{Versatile Semantic Score Map (VSSM)}. Then, by conducting common sense reasoning on \textit{VSSM}, our method can perform effective language-guided exploration and exploitation of the scene and finally reach the goal. By leveraging the reasoning and generalizing abilities of foundation models, our method can understand free-form human instructions and perform effective open-set zero-shot navigation in diverse environments. Extensive experiments on the HM3D ObjectNav benchmark show that our method surpasses all the strong baselines on all metrics, proving our method's effectiveness. Furthermore, we perform real robot demonstrations to validate our method's open-set-ness and generalizability to real-world environments.\footnote{We show further information and demo videos on \href{https://yxkryptonite.github.io/OpenFMNav/}{\texttt{https://yxkryptonite.github.io/OpenFMNav/}}.}
\end{abstract}

\section{Introduction}

As a fundamental task in robotics and embodied AI, object navigation requires an agent to navigate through unseen environments to find queried objects. Compared to other robotic tasks, it is particularly important because it is a prerequisite for robots to interact with objects. To address this issue, several household datasets and benchmarks, such as MP3D~\cite{chang2017matterport3d}, Gibson~\cite{xia2018gibson} and HM3D~\cite{hm3d} are proposed. Many previous studies~\cite{semexp, ramrakhya2022habitatweb, jiazhao} have attempted to solve this problem through supervised or reinforcement learning, where they are trained on particular household datasets above with close-set objects and comparable environments. 

\begin{figure}
\centering
\includegraphics[width=1.0\linewidth]{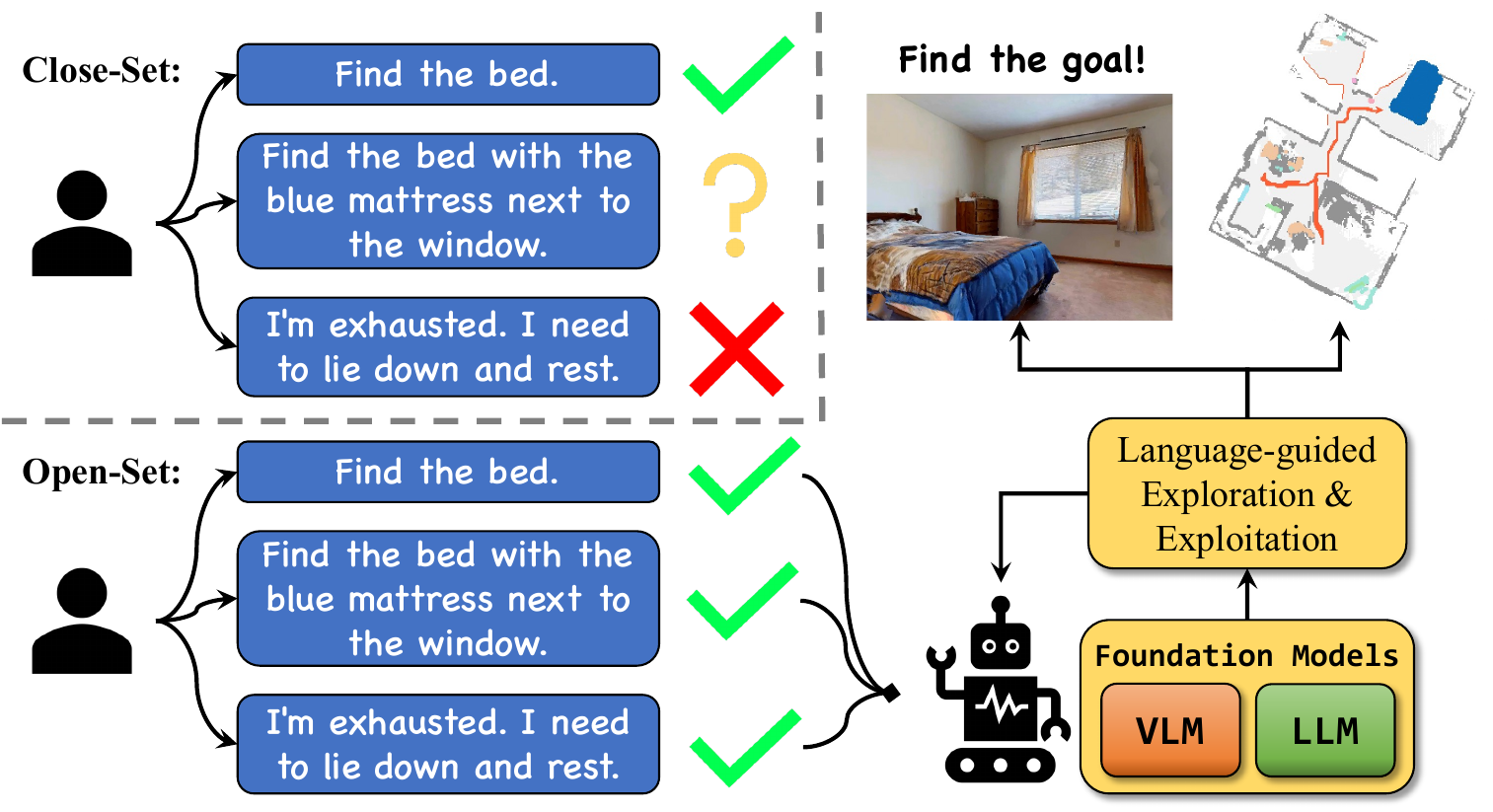}
\caption{Leveraging foundation models, our proposed OpenFMNav can follow free-form natural language instructions with open-set objects and achieve effective zero-shot object navigation.
}
\label{fig/teaser}
\end{figure}

However, there are two significant challenges remaining unsolved. \textbf{First}, as shown in Fig~\ref{fig/teaser}, in many scenarios, instead of only mentioning an object category (e.g., \emph{``Find the bed.''}), humans often provide free-form instructions, either specifying objects with specific characteristics (e.g., \emph{``Find the bed with the blue mattress next to the window.''}), or expressing their demand without explicitly mentioning the object (e.g., \emph{``I'm exhausted. I need to lie down and rest.''}). These natural language instructions may demand open-set objects not included in the training vocabulary. In such cases, existing supervised or reinforcement learning-based methods fail to understand these natural language instructions since they require specific object categories and were trained to perform close-set object detection. \textbf{Second}, due to the data scarcity of embodied navigation~\cite{survey-vln}, these methods are typically trained on limited datasets that only cover household environments, which causes severe overfitting issues and prevents them from generalizing to unseen and diverse environments, let alone performing zero-shot navigation.

To address the first challenge, some initial progress has been made in understanding free-form natural language instructions with open-set objects. For instance, demand-driven navigation (DDN) was proposed by~\citet{ddn} to map human instructions to a demand-conditioned attribute space. However, it is still limited to household settings and cannot be generalized to various environments. Another approach was suggested by~\citet{majumdar2023findthis}, which involves finding objects with specific attributes and eliminating distractors. However, it needs 2D occupancy maps and pre-exploration of the scene in the beginning, which are unavailable in unseen environments.

On the second challenge, recent years have witnessed progress in Zero-Shot Object Navigation (ZSON)~\cite{zson, cow_cvpr, yokoyama2023vlfm, zhou2023esc, lgx, yu2023l3mvn, lfg, pixnav, liang2023movln}. However, some of these works~\cite{zson, yu2023l3mvn, pixnav} require data to train specific modules such as locomotion planning, and hence are not real ``Zero-Shot''. More importantly, these methods cannot conduct explicit and comprehensive reasoning on free-form natural language instructions, leading to their low performance and preventing them from being applied to many downstream robotic tasks.

\begin{figure*}
\centering
\includegraphics[width=0.9\linewidth]{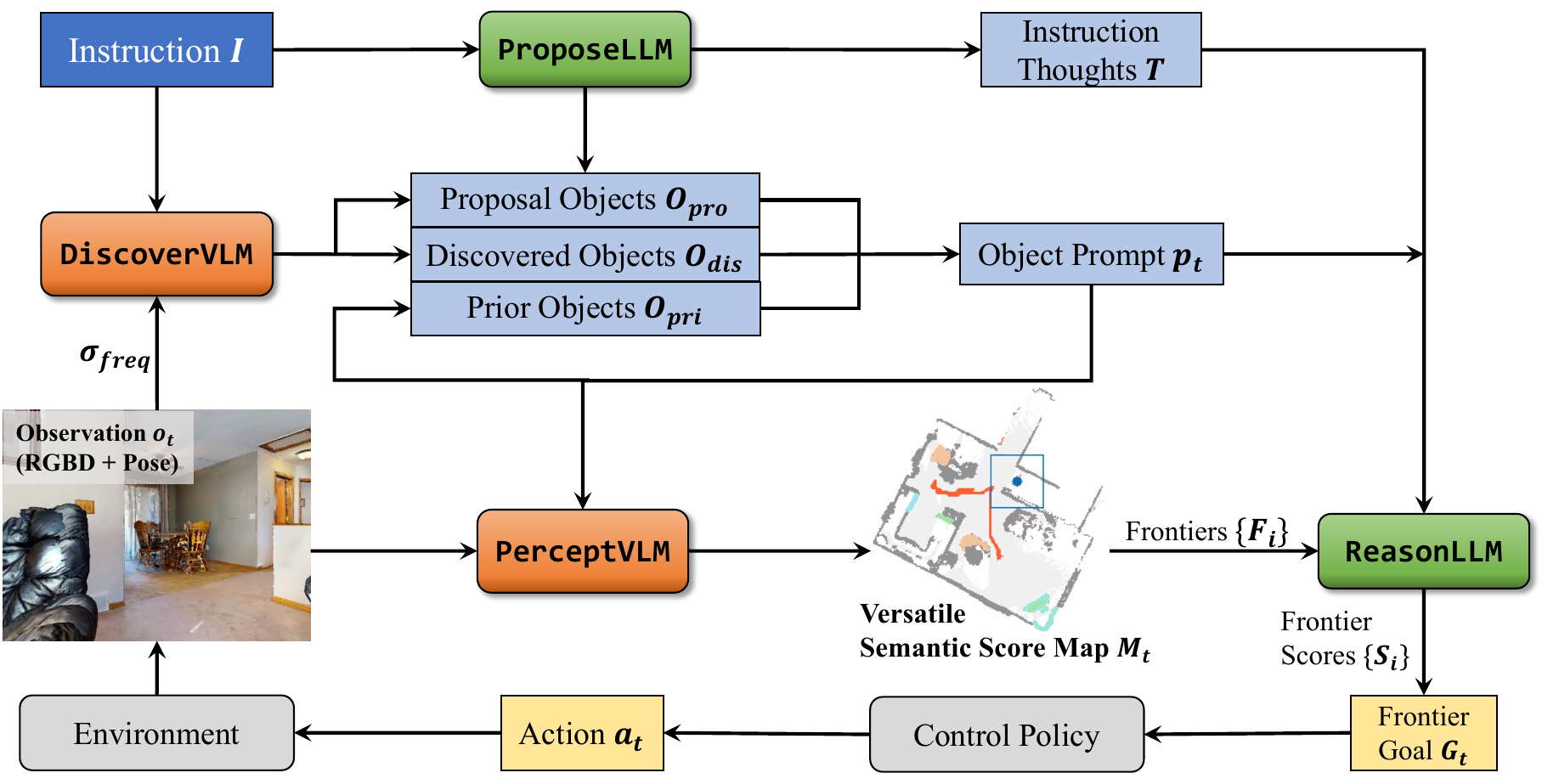}
\caption{The framework of our proposed OpenFMNav. Based on the natural language instruction and observations, we utilize foundation models to interpret human instructions and construct a \textit{Versatile Semantic Score Map (VSSM)}, on which we perform common sense reasoning and scoring to conduct language-guided frontier-based exploration.
}
\label{fig/pipeline}
\end{figure*}

To better address the aforementioned two key challenges, in this paper, we propose OpenFMNav, a novel framework based on foundation models to achieve effective open-set zero-shot navigation. To this end, we utilize foundation models to leverage their reasoning abilities and generalizability to interpret human instructions and actively explore the environment. To be more specific, we first leverage large language models to extract initially proposed objects from natural language instructions and merge them with user-defined prior objects and objects discovered by vision language models. We then construct an object prompt to detect and segment objects from the observation image, leveraging large vision language models. By using depth images to project the segmentation masks to the space, we can build a 2D top-down \textit{Versatile Semantic Score Map (VSSM)} of the whole scene, on which we sample frontiers with semantic information for a large language model to conduct common sense reasoning and wisely choose frontiers to guide navigation. This way, we can perform language-guided exploration and exploitation of the scene and achieve effective open-set zero-shot object navigation without prior training on any household datasets. Moreover, unlike previous map-based methods such as~\citet{zhou2023esc, yu2023l3mvn, lfg, yokoyama2023vlfm}, the \textit{VSSM} produced by our method will keep updating during the navigation, which better adapts to changing environments and can be further used in downstream robotic tasks, such as multi-goal navigation and mobile manipulation.

We conduct extensive experiments on the HM3D ObjectNav benchmark~\cite{habitatchallenge2022}. Results show that our method outperforms the State-of-the-Art open-set zero-shot object navigation method~\cite{zhou2023esc} by over 15\% on success rate and surpasses all the strong baselines on all metrics, validating the effectiveness and superiority of our framework. Additionally, our method has been proven to understand free-form natural language instructions with open-set objects and generalize well to real-world environments through real robot demonstrations.

\section{Related Work}

\subsection{Embodied Navigation}

Embodied navigation is a fundamental yet challenging task in robotics and embodied AI since it is the precursor to many downstream robotic tasks, such as object manipulation and teleoperation. In such scenarios, given a specific goal and egocentric observations, agents are required to move to a desired location within a maximum timestep.

Due to the importance of embodied navigation, recent years have witnessed several branches of navigation tasks with different goal specifications. For instance, point goal navigation (PointNav)~\cite{Wijmans2019DDPPOLN, savva2019habitat} uses point coordinates in the space as the goal; image goal navigation (ImageNav)~\cite{chaplot2020neural, DK18} requires the agent to move where the given image is taken; and vision-language navigation (VLN)~\cite{r2r, rxr} requires the agent to follow step-by-step instructions to reach the location; and in object navigation (ObjectNav)~\cite{batra2020objectnav}, the agent is required to find objects of specified categories.

Compared to vision-language navigation (VLN), which offers detailed and step-by-step instructions and requires an agent to strictly follow the trajectories conditioned by step-by-step instructions, object navigation (ObjectNav) is particularly challenging since the agent needs to do semantic recognition to find the goal and needs more efficient exploration than VLN since there are no step-by-step instructions~\cite{shizhe}. It is also more common in real life that humans will give ambiguous demands~\cite{ddn} rather than detailed instructions in VLN. Additionally, many VLN datasets~\cite{r2r, rxr} are typically discretized into checker-like waypoint graphs, which makes it difficult to deploy algorithms in the real world. Compared to VLN, ObjectNav is \textbf{object-centric} and \textbf{continuous} so that it can be easily deployed and extended to many downstream robotic tasks like object manipulation.

To take a step further, in this paper, we propose a solution to the problem of open-set-ness in ObjectNav by introducing a framework that transforms the paradigm of ObjectNav from given close-set category names to free-form natural language instructions with open-set objects. This transformation will help bridge the interaction between humans and embodied agents, making it more useful in real-world applications. Compared to existing works~\cite{majumdar2023findthis, ddn}, our method doesn't need prior occupancy maps and pre-exploration in the beginning and thus can navigate in unseen environments. Furthermore, our method addresses the overfitting issue in embodied navigation and easily generalizes to the real world in a zero-shot manner, enabling intelligent robot agents to navigate in more diverse environments.

\subsection{Zero-Shot Object Navigation}

As~\citet{survey-vln} elaborates, embodied navigation faces a severe challenge of data scarcity, limiting the amount and distribution of available data for training. Methods directly supervised on these limited data cannot generalize to diverse real-world environments.

Therefore, recent years have witnessed great progress in Zero-Shot Object Navigation (ZSON). Methods proposed by~\citet{zson, cow_cvpr, yokoyama2023vlfm} leverage CLIP~\cite{radford2021learning} or BLIP-2~\cite{li2023blip} embedded features to compute similarities between object goal and input image and construct an implicit map for certain goal objects to guide navigation. Other methods, such as those proposed by~\citet{zhou2023esc, lgx, yu2023l3mvn, lfg}, leverage object detectors to construct metric maps and use large language models to conduct reasoning. \citet{pixnav} leverages foundation models to perform basic image processing and trains a locomotion module to navigate to certain chosen pixel points.

\subsection{Foundation Models}\label{foundation_models}

Foundation models~\cite{bommasani2022foundation} are large-scale models that are pre-trained on vast amounts of data and can perform general tasks. The sheer volume of pretraining data endows them with exceptional generalizability, which allows them to perform zero-shot inference. Moreover, the extensive training data helps foundation models acquire common sense about our physical world, making them ideal for real-world applications.

Foundation models, particularly the large language models (LLMs), also have an intriguing feature~\textemdash~In-Context Learning (ICL)~\cite{incontext}. This feature enables these models to follow pre-defined instructions to ground their output into certain patterns. By combining ICL with common sense learned from the large-scale data, foundation models can effectively perform semantic common sense reasoning and guesswork to provide intuitions of possible exploration directions like human beings, as illustrated in~\citet{zhou2023esc, yu2023l3mvn, lfg}. For example, if the goal is a ``toilet'', from common sense it is highly possible to find it around an area that contains a ``bathtub''.

According to different modalities, foundation models can be mainly divided into Visual Foundation Models (VFM), such as SAM~\cite{kirillov2023segany}, Large Language Models (LLM), such as GPT-3.5/GPT-4~\cite{instructgpt, openai2023gpt4} and LLaMA/LLaMA-2~\cite{touvron2023llama, touvron2023llama2}, and Vision Language Models (VLM), such as GPT-4V~\cite{yang2023dawn}, CLIP~\cite{radford2021learning}, Grounded-SAM~\cite{liu2023grounding}, etc. There are also foundation models covering other modalities, such as audio~\cite{yang2023uniaudio} and video~\cite{xu2021videoclip}. In this paper, we use VLMs and LLMs since our setting only involves vision and language modalities.

\section{Method}

\subsection{Problem Statement and Method Overview}

\noindent\textbf{Problem Statement.} As shown in Fig.~\ref{fig/teaser}, in an unfamiliar environment, given a natural language instruction $I$, an embodied agent needs to explore the environment in search of a certain queried object. At timestep $t$, the agent is provided with egocentric RGBD observation $o_t$ and should output an action $a_t$ such as \texttt{move\_forward}, \texttt{turn\_left}, \texttt{stop}, etc. A successful navigation is defined as finding the queried object within the maximum navigation timestep.


\noindent\textbf{Method Overview.} As shown in Fig.~\ref{fig/pipeline}, given a starting point and human instruction $I$, the agent first utilizes the \textcolor{mydarkgreen}{ProposeLLM} to propose possible objects to meet the instruction. At timestep $t$, the agent can leverage the \textcolor{myorange}{DiscoverVLM} to discover new objects from the scene and check whether they can meet the instruction. Along with prior defined objects and proposal objects, the full object list is then converted into an object prompt $p_t$ for foundation models to reason. Given current RGBD observation $o_t$, the \textcolor{myorange}{PerceptVLM} will detect and segment object masks based on $p_t$, constructing a \textit{Versatile Semantic Score Map (VSSM)} $M_t$, on which possible exploration frontiers are sampled. Finally, the \textcolor{mydarkgreen}{ReasonLLM} will conduct common sense reasoning based on the semantic information of frontiers and give the next frontier goal $G_t$ to explore, which will be executed by an underlying control policy to output low-level actions. The whole process is looped until the object is found or the agent fails.

\subsection{Discovery and Perception}

\noindent\textbf{Discovery.} Given a free-form human instruction $I$ that may contain open-set objects, we first leverage a ProposeLLM to get all possible proposal objects $O_{pro}$ that can satisfy the instruction. Each proposal object contains attributes such as color, location, etc., to satisfy fine-grained instructions. At timestep $t$, given egocentric RGBD and pose observations $o_t$, we propose a DiscoverVLM using GPT-4V~\cite{yang2023dawn} that actively discovers novel objects $O_{dis}$ from the RGB image. Meanwhile, the DiscoverVLM also conducts reasoning on the instruction, trying to discover objects that potentially meet the instruction and update $O_{pro}$. Extracting novel objects from the environment is essential for open-set navigation since they may contain scene-specific information that helps to find the goal. To save time and cost, the DiscoverVLM is randomly activated by a frequency parameter $\sigma_{freq}$.

\noindent\textbf{Perception.} After getting proposal objects $O_{pro}$ and discovered objects $O_{dis}$, we merge them with prior objects $O_{pri}$ to construct an object prompt $p_t$ to feed into our PerceptVLM based on Grounded-SAM~\cite{liu2023grounding} to detect and segment all the appearing objects in $p_t$ from the RGB image of $o_t$. Note that due to the BERT encoder~\cite{devlin-etal-2019-bert} and powerful SAM backbone~\cite{kirillov2023segany} in the PerceptVLM, it can achieve open-set object detection in high granularities. This process will output object masks with confidence scores for further mapping and reasoning.

\subsection{Mapping and Reasoning}

\noindent\textbf{Mapping.} At timestep $t$, based on the confidence scores of object masks produced by PerceptVLM and the depth image and pose in $o_t$, we project the masks to the top-down 2D space and construct a \textit{Versatile Semantic Score Map (VSSM)} $M_t \in \mathbb{R}^{H\times W\times (C+2)}$, which contains $C$ channels of object semantics, and two channels of the occupied area and explored area, with a resolution of $H\times W$. Each element in the map is a score in $[0,1]$ instead of binary labels. Since we continuously discover novel objects from the environment, the $C$ is versatile so that we can keep updating the map, enabling life-long learning and downstream robotic tasks. Also, instead of filling binary labels into semantic channels, we fill each semantic channel with confidence scores, with which we can easily update the map if there is a change in the environment.

\noindent\textbf{Reasoning.} Based on $M_t$, we can sample frontiers $\{F_i\}$ with semantic information in unexplored areas for further exploration. To choose the next frontier to explore, we leverage ReasonLLM by unleashing the power of LLM's common sense reasoning. Specifically, given the semantic information around each frontier, we construct a query template in the form of \texttt{``This area contains A, B and C.''}. Combined with the thought $T$ produced by Chain-of-Thought~\cite{wei2022chain} prompting from ProposeLLM and the object prompt $p_t$, the ReasonLLM will conduct common sense reasoning as in Section~\ref{foundation_models} and rate these frontiers to pick one frontier goal $G_t$ which is most likely to find the object goal. This frontier goal $G_t$ will guide the agent for further exploration and produce low-level actions to control the agent.

Instead of directly asking the LLM which frontier to explore for once or multiple times~\cite{lfg}, we leverage another reasoning process, which prompts the LLM to rate these frontiers $\{F_i\}$ to scores $\{S_i\}$, in which $S_i \in [0,1]$, indicating the likelihood to find the goal. Then, the frontier with the highest score will be picked out for further exploration. By leveraging this rating process, ReasonLLM can map its common sense to concrete numbers that reflect the actual ranking, leading to better reasoning. We verified its effectiveness in Section~\ref{ablation}. It's also worth mentioning that to balance exploration and exploitation, ReasonLLM is activated at regular timestep intervals $\delta$ to update $G_t$. At other timesteps, the frontier goal $G_t$ remains unchanged to fully explore the previously chosen frontier $G_{t-\delta}$.

After obtaining the frontier goal and the occupancy channel in $M_t$, we utilize a control policy based on the Fast Marching Method (FMM)~\cite{sethian1999fast} to output a low-level action $a_t$ to control the agent. This closes the loop and goes to the next timestep $t+1$.

We present the whole process of our OpenFMNav algorithm in Algorithm~\ref{algo}.

\begin{algorithm}[t]\small
\SetAlgoLined
\caption{Pseudo-Code of the Overall Algorithm for OpenFMNav}\label{algo}
\KwData{Natural Language Instruction \(I\), Prior Objects \(O_{pri}\), Discovery Frequency \(\sigma_{freq}\), Frontier Goal Update Interval \(\delta\)}
\(t \gets 0\)\;
\(done \gets False\)\;
\(G_0,~M_0,~O_{dis} \gets None\)\;
\(O_{pro}, T\gets \) \textcolor{mydarkgreen}{ProposeLLM}(\(I\))\;
\While{not done}{
    \(o_t \gets \) getObservation()\;
    \If{toDiscover(\(\sigma_{freq}\))}{
        \(O_{dis},~O_{pro }\gets \) \textcolor{myorange}{DiscoverVLM}(\(o_t,~I\))\;
    }
    \(p_t \gets \) getPrompt(\(O_{pro},~O_{dis},~O_{pri}\))\;
    \(Masks \gets \) \textcolor{myorange}{PerceptVLM}(\(o_t,~p_t\))\;
    \(M_t \gets \) semanticMapping(\(M_{t-1},~Masks,~o_t\))\;
    \uIf{\(O_{pro}\) in \(M_t\)}{
        \(G_t \gets \) getLocation(\(M_t,~O_{pro}\))\;
    }
    \Else{
        \uIf{\(t~\%~\delta == 0\)}{
            \(\{F_i\} \gets \) sampleFrontiers(\(M_t\))\;
            \(\{S_i\} \gets \) \textcolor{mydarkgreen}{ReasonLLM}(\(\{F_i\},~p_t,~T\))\;
            \(G_t \gets \) getLocation(\(M_t\),~argmax(\(\{S_i\}\)))\;
        }
        \Else{
            \(G_t \gets G_{t-1}\)\;
        }
    }
    \(O_{pri} \gets \) updateObj(\(O_{pro},~O_{dis},~O_{pri}\))\;
    \(a_t \gets \) FMMPlanner(\(M_t,~G_t\))\;
    \(done \gets \) stepAction(\(a_t,~t\))\;
    \(t \gets t+1\)\;
}
\end{algorithm}

\section{Experiments}

\begin{table*}
\centering
\begin{tabular}{ccccc}
\hline
\textbf{Method} & \textbf{Open-Set} & \textbf{Zero-Shot} & \textbf{SR (\%) $\uparrow$} & \textbf{SPL $\uparrow$}\\
\hline
FBE~\citep{FBE} & $\times$ & \checkmark & 23.7 & 0.123 \\
SemExp~\citep{semexp} & $\times$ & $\times$ & 37.9 & 0.188 \\
ZSON~\citep{zson}  & \checkmark & $\times$ & 25.5 & 0.126 \\ 
GoW~\citep{cow_cvpr} & \checkmark & \checkmark & 32.0 & 0.181 \\ 
ESC~\citep{zhou2023esc} & \checkmark & \checkmark & 38.5 & 0.220 \\ 
L3MVN~\citep{yu2023l3mvn} & $\times$ & \checkmark & 50.4 & 0.231 \\ 
L3MVN + GPT-4~\cite{yu2023l3mvn} & $\times$ & \checkmark & 51.8 & 0.234 \\
PixNav~\citep{pixnav} & $\checkmark$ & $\times$ & 37.9 & 0.205 \\
\hline
OpenFMNav (Ours) & \checkmark & \checkmark & \textbf{54.9} & \textbf{0.244} \\ \hline
\end{tabular}
\caption{Comparison between different methods on the HM3D ObjectNav benchmark. Our method outperforms all the baseline methods on all metrics and achieves open-set zero-shot object navigation.}
\label{tab/comparison}
\end{table*}

In this section, we evaluate our method comprehensively in simulation to show our method's effectiveness compared to baseline methods. We also conducted ablation studies to validate the effectiveness of our framework design.

\subsection{Experimental Setup}

In the simulation, we evaluate on the HM3D ObjectNav benchmark based on the Habitat Matterport 3D Semantics Dataset~\cite{hm3dsem}, which contains 80 train scenes and 20 validation scenes. We utilize the validation scenes for evaluation. There are, in total, 2000 episodes and six goal classes (chair, couch, potted plant, bed, toilet, and tv) in the dataset. The action space of the robot agent is \texttt{\{stop, move\_forward, turn\_left, turn\_right, look\_up, look\_down\}}. The forward distance is set to 0.25m, and the rotation angle is set to 30 degrees.

Following previous works~\cite{zhou2023esc, pixnav}, we utilize \textbf{Success Rate (SR)} metric to measure whether an agent can find our desired objects. We also report results of \textbf{Success weighted by Path Length (SPL)} to measure the navigation efficiency.

\subsection{Implementation Details}

In our method, the foundation models we use are: GPT-4 (text-only)~\cite{openai2023gpt4} for ProposeLLM and ReasonLLM, and GPT-4V~\cite{yang2023dawn} for DiscoverVLM. For PerceptVLM, we utilize Grounded-SAM, which first leverages Grounding DINO~\cite{liu2023grounding} to produce bounding boxes given the RGB image in $o_t$ and object prompt $p_t$, and then leverages Segment Anything Model (SAM)~\cite{kirillov2023segany} for each bounding box to produce high-granularity object masks for semantic mapping.

Moreover, we utilize the Chain-of-Thought (CoT)~\cite{wei2022chain} prompting technique to fully exploit the reasoning abilities of ProposeLLM, ReasonLLM and DiscoverVLM. The prompts we used can be found in Appendix~\ref{prompts}.

In the simulation, we set the update interval $\delta$ to 20 timesteps, discovery frequency $\sigma_{freq}$ to 0.01, and the initial prior objects to a subset of HM3D object categories, which can be found in Appendix~\ref{hyper}.

\subsection{Baseline Methods}

We compare our method with several recent works, with a focus on open-set and zero-shot object navigation baselines to verify our framework's effectiveness. We classify these baseline methods into ``Open-Set'' and ``Zero-Shot'' or not. Here, we define ``Open-Set'' as that the method can find whatever object category we want, and define ``Zero-Shot'' as that the agent hasn't been trained or finetuned on \textbf{\textit{any}} of the data previously, including images, episodes, and locomotion planning. The baseline methods are as follows:


\begin{itemize}
    \item \textbf{FBE}~\cite{FBE}. This baseline method employs a classical robotics pipeline for mapping and a frontier-based exploration algorithm.
    \item \textbf{SemExp}~\cite{semexp}. A method that explores and searches for the target using close-set semantic maps and reinforcement learning.
    \item \textbf{ZSON}~\cite{zson}. An RGB-based zero-shot object navigation baseline using CLIP~\cite{radford2021learning} to embed scene features. It is trained on ImageNav and directly transferred to ObjectNav.
    \item \textbf{GoW}~\cite{cow_cvpr}. A modification of CoW~\cite{cow_cvpr} implemented by~\citet{zhou2023esc} that uses GLIP~\cite{glip} for object detection and the vanilla fronter-based exploration method.
    \item \textbf{ESC}~\cite{zhou2023esc}. A map-based zero-shot object navigation baseline that uses GLIP~\cite{glip} to detect objects and rooms, and combines LLM with soft commonsense constraints for planning.
    \item \textbf{L3MVN}~\cite{yu2023l3mvn}. An LLM-based baseline that finetunes a close-set object detector~\cite{jiang2018rednet} and an LLM to conduct frontier-based exploration. We also conduct experiments that replace its LLM with GPT-4 for fairer comparisons.
    \item \textbf{PixNav}~\cite{pixnav}. A recent work that solely uses foundation models to pick out navigation pixels and trains a locomotion module to navigate to chosen pixels.
\end{itemize}

\begin{table}
\centering
\begin{tabular}{ccc}
\hline
\textbf{Method} & \textbf{SR (\%) $\uparrow$} & \textbf{SPL $\uparrow$}\\
\hline
w/o GPT-4 & 53.6 & 0.230 \\
w/o CoT & 51.8 & 0.208 \\
w/o Discovery & 50.0 & 0.222 \\
w/o Scoring & 50.0 & 0.208 \\
\hline
Ours & \textbf{55.4} & \textbf{0.239} \\ \hline
\end{tabular}
\caption{Ablation studies on different components of our method. Experiments are conducted under the same uniformly sampled episodes.}
\label{tab/ablation}
\end{table}

\subsection{Results and Analysis}

We report the main results in Table~\ref{tab/comparison}. Our method surpasses all the baselines on both Success Rate (SR) and Success weighted by Path Length (SPL), especially compared with open-set zero-shot methods. Our method surpasses the previous State-of-the-Art method on open-set zero-shot object navigation~\cite{zhou2023esc} by over 15\% on the success rate metric, suggesting that our framework is indeed effective.

First, we compare our method with previous semantic map based methods, such as SemExp~\cite{semexp}, ESC~\cite{zhou2023esc} and L3MVN~\cite{yu2023l3mvn}. The results show that our method performs better since we utilize DiscoverVLM to construct \textit{VSSM} with versatile out-of-vocabulary class labels, such as ``marble statue'' and ``range hood'', which helps to alleviate the issue of limited categories and enriches the semantic information of the environment. Also, compared to these methods, our method achieves open-set navigation, which better adapts to complex situations and real-world applications.

Compared with other open-set baselines, such as PixNav~\cite{pixnav}, ZSON~\cite{zson} and GoW~\cite{cow_cvpr}, our method constructs an explicit map where all discovered objects are presented. Therefore, we can boost LLMs' reasoning abilities to balance between exploration and exploitation and make the agent move to where the goal is most likely to be. Also, the map constructed by our method is maintained and updated, which is perfect for life-long learning, enabling downstream robotic tasks with further natural language instructions, while methods like \citet{cow_cvpr, yokoyama2023vlfm} only construct implicit maps for a certain goal, which is useless after the navigation.

\begin{figure}
\centering
\includegraphics[width=\linewidth]{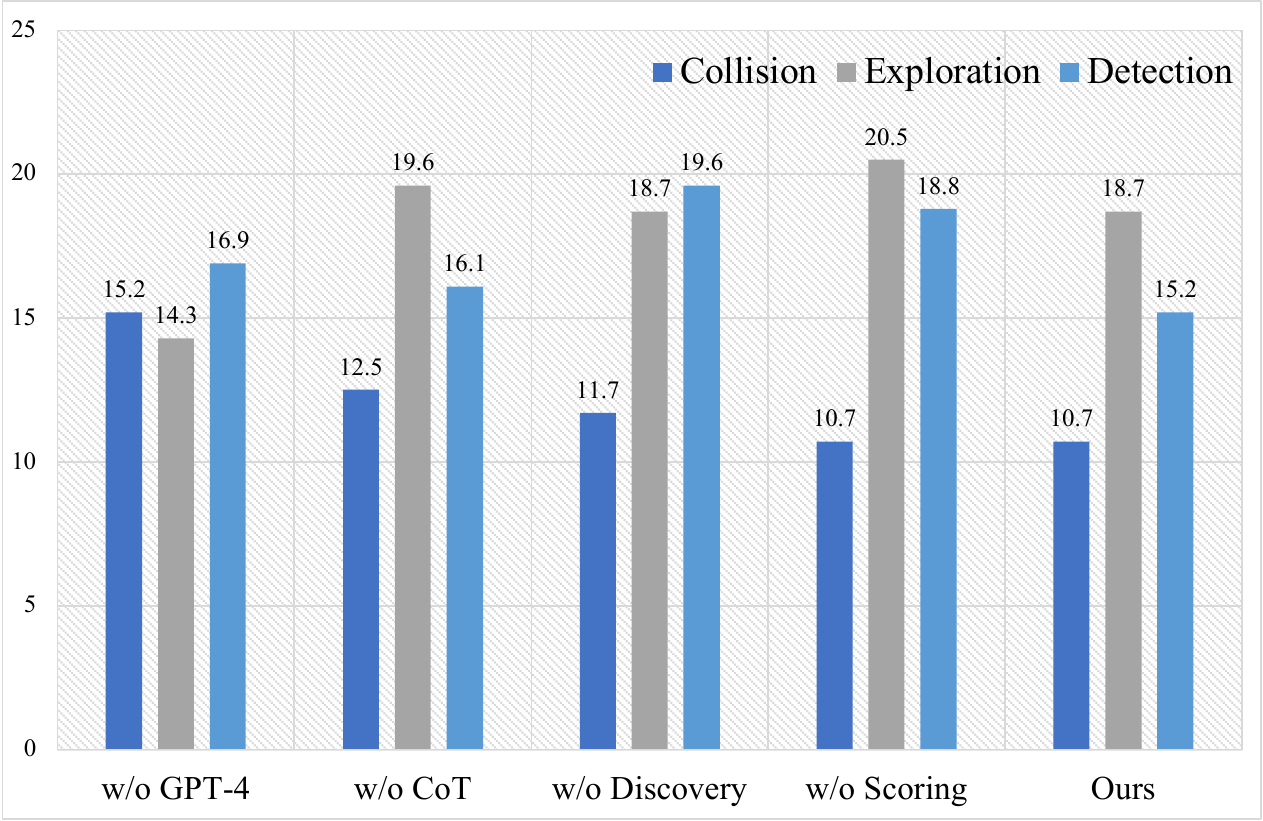}
\caption{Types and percentages of failure cases in ablation methods.
}
\label{fig/chart}
\end{figure}


\begin{figure*}
\centering
\includegraphics[width=\linewidth]{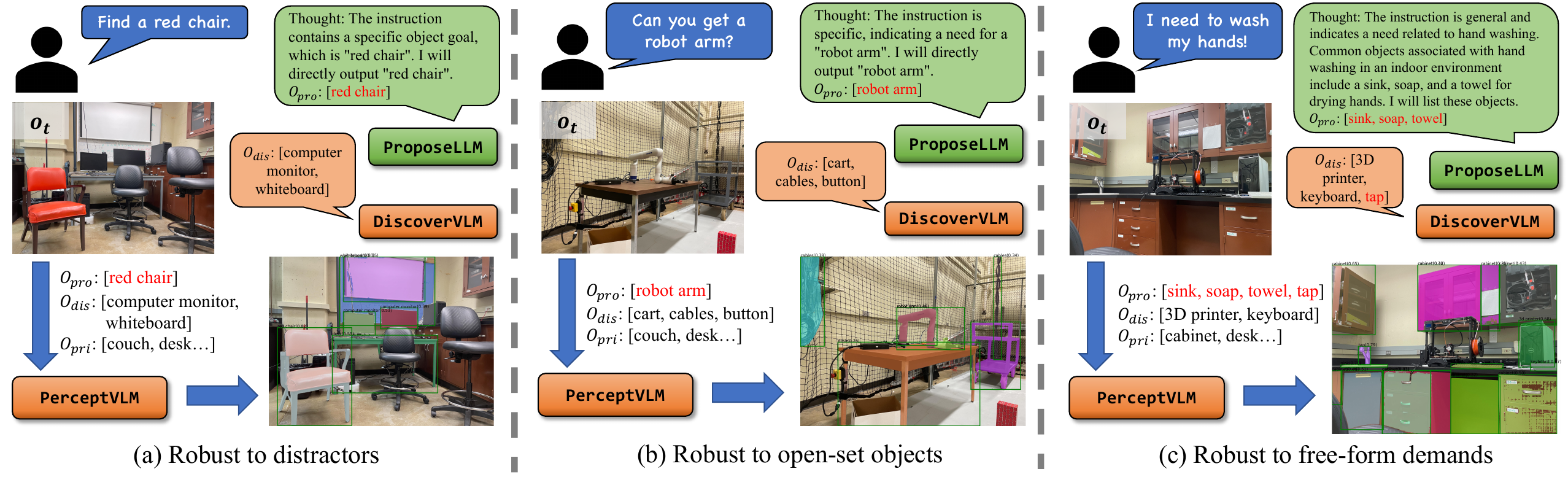}
\caption{Qualitative studies in the real world. Text marked in \textcolor{red}{red} indicates objects that potentially satisfy the instruction. Results show that our method is robust to natural language instructions, including distractors, open-set objects and free-form demands.
}
\label{fig/threecase}
\end{figure*}

\subsection{Ablation Studies}\label{ablation}

Probing deeper into our method design, we also performed ablation studies on various components of our pipeline. Note that to save time and cost, we test all the ablation methods on a subset of the full dataset under the same uniformly sampled episodes so that there can be slight differences in the result of our method. Table~\ref{tab/ablation} shows that modifying multiple components of our framework leads to significantly worse performance. We also categorized the failure cases into different types and report their percentages in Fig.~\ref{fig/chart}, in which \textbf{Collision} refers to the situation where the agent cannot avoid colliding with the environment, \textbf{Exploration} means the agent times out while trying to find the goal, and \textbf{Detection} means the agent mistakenly identifies a wrong object as the goal.

\noindent\textbf{Effectiveness of using larger models.} First, we analyze the usage of GPT-4 for LLMs. Compared to only using GPT-3.5, using larger GPT-4 achieves better performance ($+1.8\%$), reducing failure cases of Collision and Detection. However, the percentage of Exploration is slightly higher, showing that larger models have more diverse answers that encourage more exploration, which potentially causes more time out.

\noindent\textbf{Effectiveness of our joint reasoning pipeline.} Then, we analyze different foundation model components. We found that using CoT prompting ($+3.6\%$) and scoring prompting ($+5.4\%$) are essential to the strong performance of OpenFMNav since they generate more reasoning chains that elicit the common sense of large language models. Also, compared to restricting the object set, leveraging DiscoverVLM not only enables more free-form natural language instructions from users' input but also enriches the scene's semantics, which helps the reasoning for frontier-based exploration and improves performance ($+5.4\%$). These efforts reduce failure cases of all categories.

\section{Navigation in the Real World}

We further conduct real robot demonstrations to show our method's ability to understand free-form natural language instructions and perform open-set zero-shot navigation in the real world.

\subsection{Real Robot Setup}

For robots, we use a TurtleBot4 robot with scalable structures to navigate on the ground. We limit its action space to \texttt{\{stop, move\_forward, turn\_left, turn\_right\}}. As in the simulation, we set the forward distance to 0.25m and the rotation angle to 30 degrees. For robotic perception, we use a Kinect RGBD camera to capture RGBD images.

For real-world environments, we select multiple rooms (including offices, labs, and meeting rooms) with sufficient space and various objects for the robot to navigate. These rooms contain not only common objects like ``chair'', ``couch'', ``desk'', ``computer'', ``cabinet'', etc., but also less common ones like ``robot arm'', ``3D printer'', ``coffee machines'', etc.


\subsection{Qualitative Studies}

We conduct qualitative studies on our OpenFMNav in the real world, as shown in Fig.~\ref{fig/threecase}. The results show that our method can perform effective zero-shot navigation in the real world given free-form natural language instructions. Especially, our method is robust to distractors, open-set objects and free-form demands.

For distractors, rather than object categories, our proposed ProposeLLM can extract the attributes in the instruction (``red chair''), which can be further detected and segmented by PerceptVLM. In Fig.~\ref{fig/threecase}(a), we can see that, among the three chairs in the observation, only the red chair is masked.

For open-set objects, due to the large-scale training data of foundation models, our method can also navigate to objects that are uncommon and out-of-vocabulary, such as the ``robot arm'' in Fig.~\ref{fig/threecase}(b).

Another intriguing feature of our method is that our method can adaptively add up goals in the navigation. This happens when the instruction is a free-form demand for ambiguous objects. For example, in Fig.~\ref{fig/threecase}(c), when the user needs to wash hands, the ProposeLLM first proposed ``sink'', ``soap'' and ``towel'', but they are not necessarily present in the scene. When the agent explores the environment, the DiscoverVLM can actively discover what's new in the environment and reason about whether they can potentially fulfill the user's demand. In this case, a ``tap'' is discovered and identified as a goal so that the agent can directly navigate to it without further exploration. This is extremely helpful when the humans are also unaware of the scene details.

\section{Conclusions}

In this paper, we presented a novel framework, OpenFMNav, for open-set zero-shot object navigation. By leveraging foundation models, our method could understand free-form natural language instructions, conduct reasoning, and perform effective zero-shot object navigation. Extensive experiments showed the superiority of our framework. Finally, we conducted real robot demonstrations to validate our method's open-set-ness and generalizability to real-world environments.

\section*{Ethics Statement}

In this paper, we present a method for open-set zero-shot object navigation. This method can be used for zero-shot robotic navigation in diverse scenarios, such as home robots, warehouse robots, and so on. Our work further addresses the issue of ambiguous or free-form natural language instructions, benefitting the interaction between humans and robots. However, foundation models can have safety issues and risks such as privacy leaks and jailbreaking~\cite{deng2023masterkey, chao2023jailbreaking}, which need to be further addressed.

\section*{Limitations}

While extensive experiments validate the effectiveness of our method design, there exist a number of limitations in our work. First, our method requires relatively accurate depth sensors to build the 2D map, while the observed depths and camera poses may have much noise in reality, causing performance degradation. Moreover, we acknowledge that our method requires stable Internet connections to get responses from APIs of foundation models, limiting the potential of large-scale deployment in harsh environments. Another limitation is that the use of LLMs may not always be real-time, which can cause latency issues. We hope future works on depth sensing, LLM quantization, and edge computing can mitigate such limitations.

\section*{Acknowledgements}

This work was partially supported by NSF IIS-2119531, IIS-2137396, IIS-2142827, IIS-2234058, CCF-1901059, and ONR N00014-22-1-2507.

\bibliography{custom}
\newpage

\appendix

\section{API Usage}\label{api}

\begin{table}[h]
\centering
\begin{tabular}{cc}
\hline
\textbf{Model Name} & \textbf{API Name} \\
\hline
\textcolor{mydarkgreen}{ProposeLLM} & \texttt{gpt-4-1106-preview} \\ 
\textcolor{myorange}{DiscoverVLM} & \texttt{gpt-4-vision-preview} \\ 
\textcolor{mydarkgreen}{ReasonLLM} & \texttt{gpt-4-1106-preview} \\
\hline
\end{tabular}
\caption{API usage}
\label{tab/api}
\vspace{-15pt}
\end{table}

\section{Hyperparameters}\label{hyper}

\begin{table}[h]
\centering
\begin{tabular}{cc}
\hline
\textbf{Parameter} & \textbf{Value} \\
\hline
Discovery Frequency $\sigma_{freq}$ & 0.01 \\
Frontier Goal Update Interval $\delta$ & 20 \\
Confidence Score Threshold & 0.55 \\
LLM/VLM Temperature & 0 \\
Initial Prior Objects $O_{pri}$ & See Fig.~\ref{fig/obj} \\
\hline
\end{tabular}
\caption{Hyperparameters}
\label{tab/hyperparameter}
\vspace{-15pt}
\end{table}

\begin{figure}[ht]
  \centering
  \framebox{\parbox{2.5in}{
  chair, bed, plant, toilet, tv, couch, desk, refrigerator, sink, bathtub, shower, towel, painting, trashcan, stairs
  }}
  \caption{Initial prior objects $O_{pri}$}
  \label{fig/obj}
  \vspace{-15pt}
\end{figure}

\section{Prompts and Examples}\label{prompts}

Below we show prompts and examples of LLM input/output.

\subsection{Prompts for ProposeLLM}

\begin{figure}[!h]
\centering
\includegraphics[width=\linewidth]{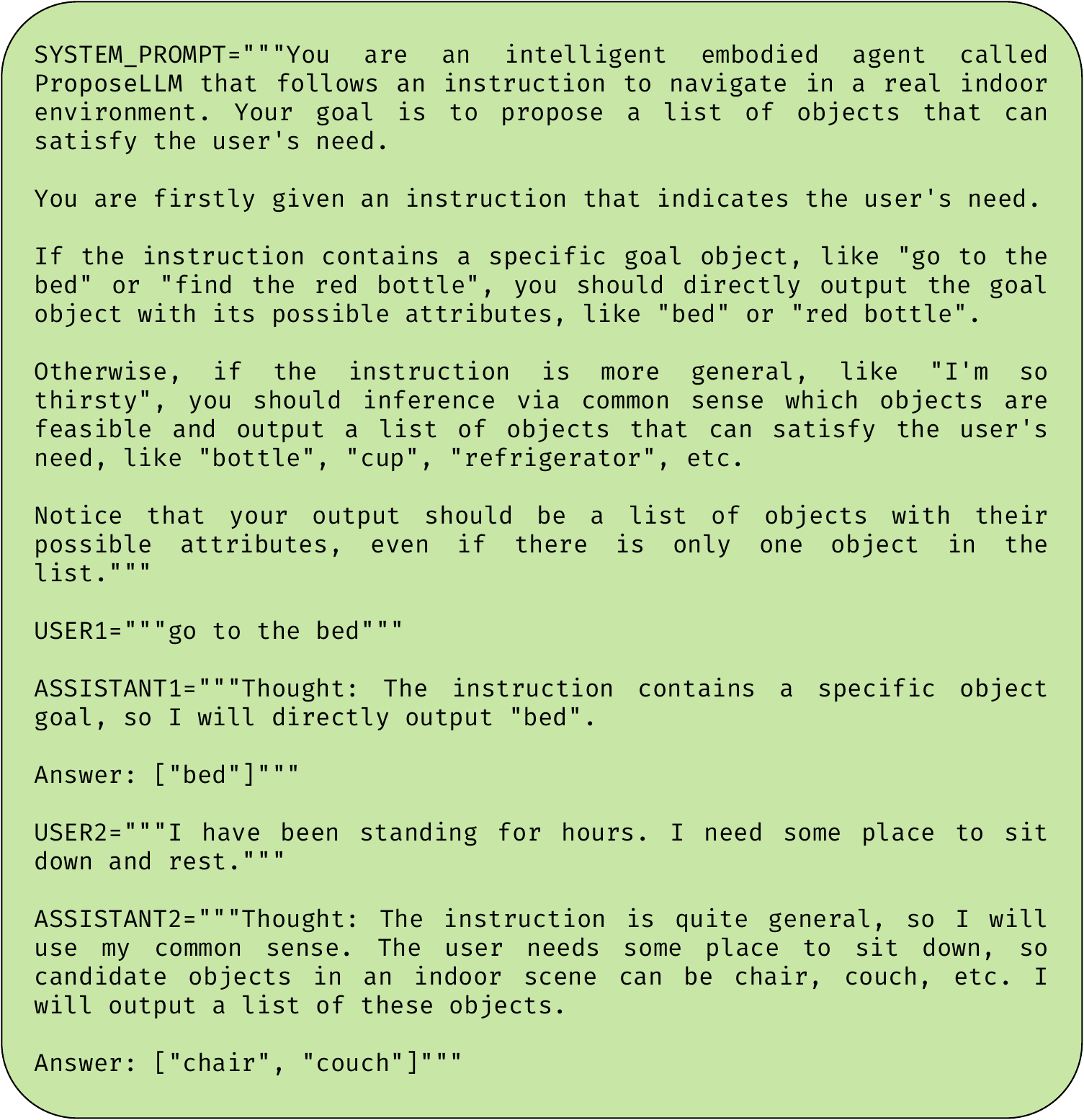}
\caption{Prompts for ProposeLLM}
\label{fig/propose}
\end{figure}

The prompts for ReasonLLM are shown in Fig.~\ref{fig/propose}.

\subsection{Prompts for DiscoverVLM}

\begin{figure}[ht]
\centering
\includegraphics[width=\linewidth]{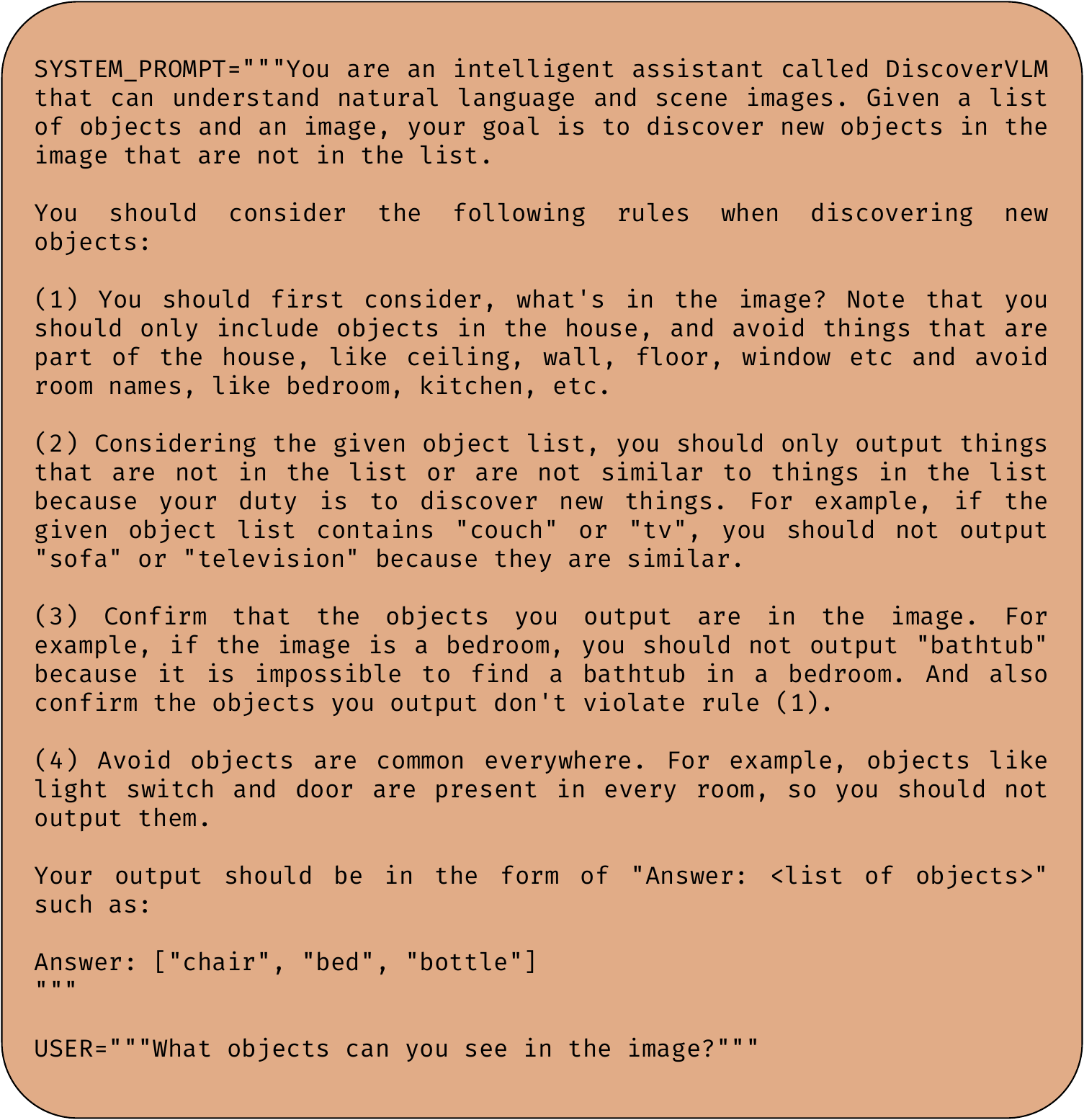}
\caption{Prompts for DiscoverVLM}
\label{fig/discover}
\end{figure}

The prompts for ReasonLLM are shown in Fig.~\ref{fig/discover}.

\subsection{Prompts for PerceptVLM}

For PerceptVLM, given the current object list, we use dots to separate each object as the object prompt $p_t$.

For example, if the object list is \texttt{[chair, bed, plant, toilet, tv, couch]}, the object prompt is \texttt{``chair.bed.plant.toilet.tv.couch''}.

\subsection{Prompts for ReasonLLM}

\begin{figure*}[ht]
\centering
\includegraphics[width=\linewidth]{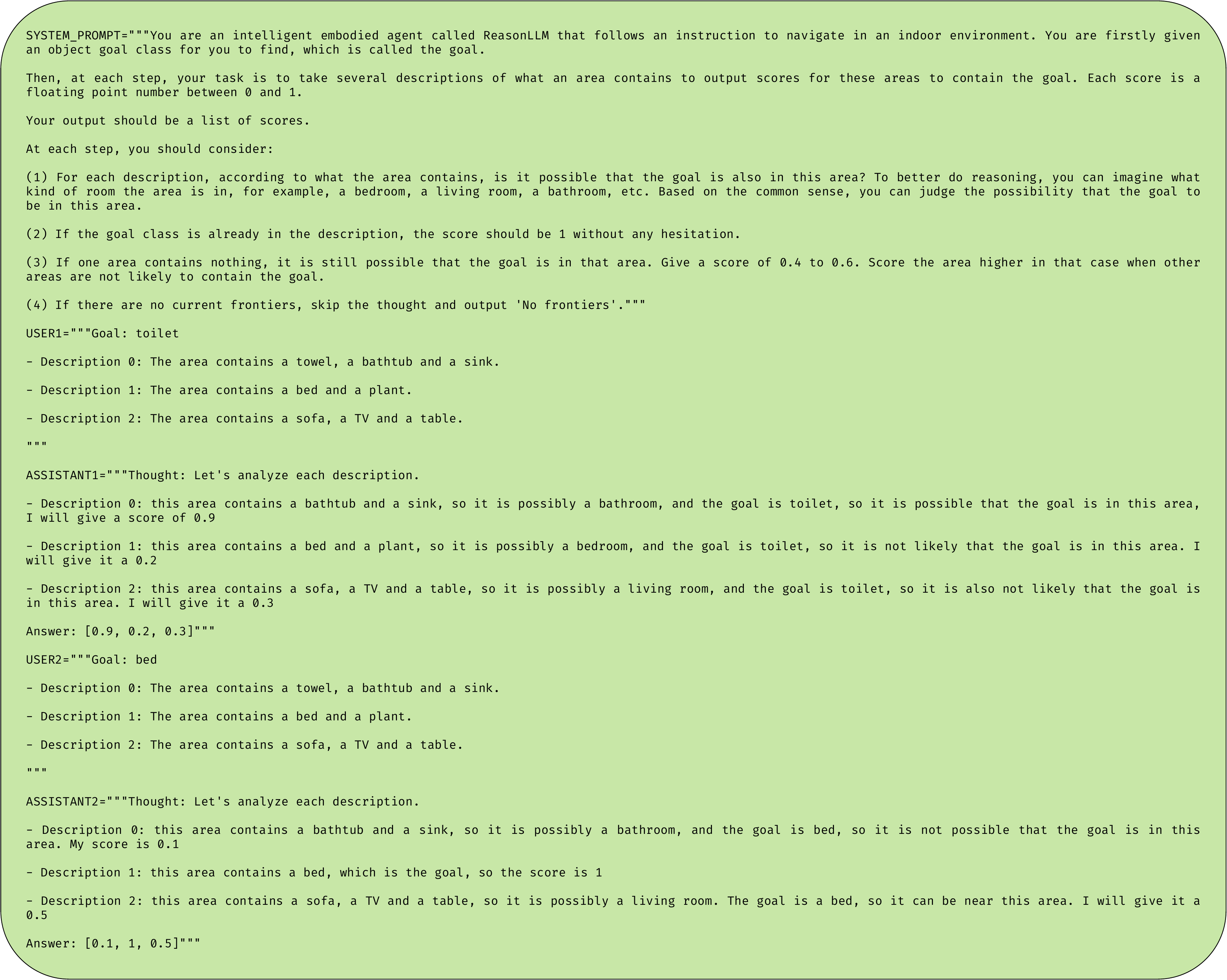}
\caption{Prompts for ReasonLLM}
\label{fig/reason}
\end{figure*}

The prompts for ReasonLLM are shown in Fig.~\ref{fig/reason}.

\end{document}